\documentclass[10pt,twocolumn,letterpaper]{article}

\usepackage{cvpr}              % To produce the CAMERA-READY version
\usepackage{graphicx}
\usepackage{amsmath}
\usepackage{amssymb}
\usepackage{booktabs}
\usepackage{algorithm}
\usepackage{algpseudocode}
\usepackage{graphics}
\usepackage{times}
\usepackage{epsfig}
\usepackage{float}
\usepackage{multirow}
\usepackage{enumitem}
\usepackage{graphicx, amsmath, amssymb, caption, subcaption, multirow, overpic, textpos}
\usepackage[table,dvipsnames]{xcolor}
\usepackage[british, english, american]{babel}
\definecolor{citecolor}{HTML}{0071BC}
\definecolor{linkcolor}{HTML}{ED1C24}
\usepackage[colorlinks,linkcolor=blue]{hyperref}
\def\cvprPaperID{**}
\def\confName{****}
\def\confYear{****}

\makeatletter
\DeclareRobustCommand\onedot{\futurelet\@let@token\@onedot}
\def\@onedot{\ifx\@let@token.\else.\null\fi\xspace}

\def\ie{\emph{i.e}\onedot}

\makeatother

\newlength\savewidth

\renewcommand{\paragraph}[1]{\vspace{1.25mm}\noindent\textbf{#1}}

\newcolumntype{x}[1]{>{\centering\arraybackslash}p{#1pt}}
\newcolumntype{y}[1]{>{\raggedright\arraybackslash}p{#1pt}}
\newcolumntype{z}[1]{>{\raggedleft\arraybackslash}p{#1pt}}

\newcommand{\app}{\raise.17ex\hbox{$\scriptstyle\sim$}}

\definecolor{deemph}{gray}{0.6}

\definecolor{baselinecolor}{gray}{.9}

\begin{document}
	
	%%%%%%%%% TITLE
	\title{MetaDance: Few-shot Dancing Video Retargeting \\via Temporal-aware Meta-learning}
	\author{Yuying Ge$^1$ \quad
	Yibing Song$^{2}$ \quad 
	Ruimao Zhang$^{3,4}$ \quad
	Ping Luo$^1$\\
	{$^1$The University of Hong Kong} \quad \quad {$^2$Tencent AI Lab} \\ {$^3$The Chinese University of Hong Kong (Shenzhen)}\\
	{$^4$Shenzhen Research Institute of Big Data}\\
}

	\maketitle
		
%%%%%%%%% ABSTRACT
\vspace{-25pt}
\begin{abstract}
\vspace{-5pt}
Dancing video retargeting aims to synthesize a video that transfers the dance movements from a source video to a target person. Previous work need collect a several-minute-long video of a target person with thousands of frames to train a personalized model. However, the trained model can only generate videos of the same person. To address the limitations, recent work tackled few-shot dancing video retargeting, which learns to synthesize videos of unseen persons by leveraging a few frames of them. In practice, given a few frames of a person, these work simply regarded them as a batch of individual images without temporal correlations, thus generating temporally incoherent dancing videos of low visual quality. In this work, we model a few frames of a person as a series of dancing moves, where each move contains two consecutive frames, to extract the appearance patterns and the temporal dynamics of this person. We propose MetaDance, which utilizes temporal-aware meta-learning to optimize the initialization of a model through the synthesis of dancing moves, such that the meta-trained model can be efficiently tuned towards enhanced visual quality and strengthened temporal stability for unseen persons with a few frames. Extensive evaluations show large superiority of our method.
\end{abstract}

%%%%%%%%% BODY TEXT
\section{Introduction}
\begin{figure}[t]
	\begin{center}
		\includegraphics[width=\linewidth]{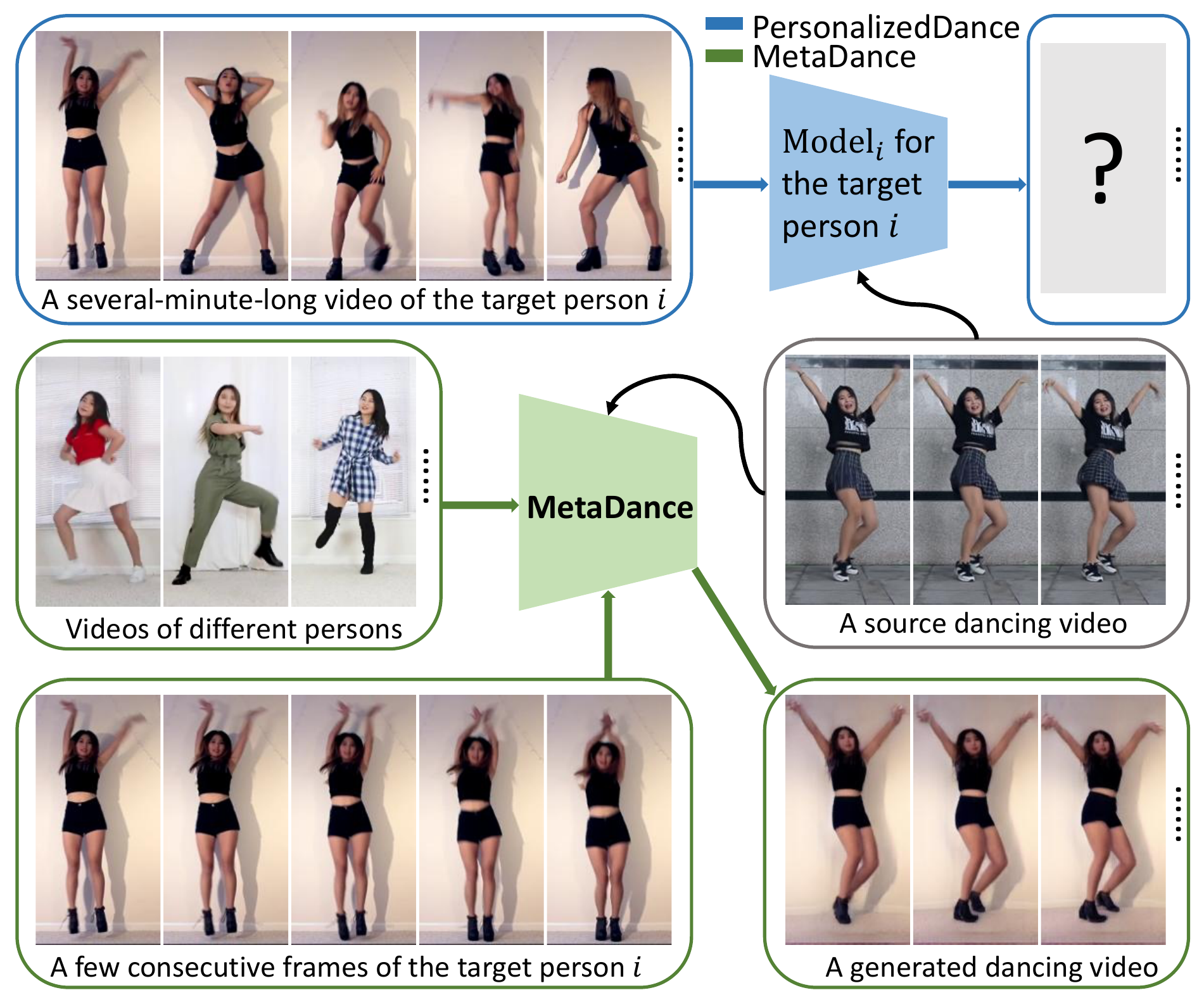}
	\end{center}
	\vspace{-10pt}
	\caption{The comparison between previous work~\cite{vid2vid,everybody,dancegen,2021supervised} referred as PersonalizedDance and our MetaDance in dancing video retargeting. To retarget a source dancing video for a new target person, PersonalizedDance need collect a several-minute long video of this person with thousands of frames to train a personalized model. By contrast, MetaDance only need a few frames of this person to tune a generalized model, which is obtained after being trained on dancing videos of different persons.} 
	\vspace{-10pt}
	\label{fig:few-shot}
\end{figure}

Dancing video retargeting aims to transfer the dance movements from a source video to a target person, thus synthesizing a video where the target person mimics the dancer in the source video as shown in Fig~\ref{fig:few-shot}. This task has great potential in application because it enables everybody to dance like professional dancers in videos. Since a dancer in a video quickly changes postures with large movements, it is extremely challenging to synthesize a high-quality video for a target person performing the same dance.
In order to retarget a source dancing video for a target person, previous methods~\cite{vid2vid,everybody,dancegen,2021supervised} referred as PersonalizedDance first collected a several-minute-long video of this person performing moves (\ie 3 $\sim$ 4 minutes long for a target person in vid2vid~\cite{vid2vid}). The collected video contains thousandths of frames, which capture a sufficient range of motions of the target person.
Then these methods trained a personalized model for the target person, which can only synthesize dancing videos for this specific person as shown in Fig~\ref{fig:few-shot}. Since these methods lacked the generalization ability, adopting them to retarget a dancing video for a new person requires collecting another video of the corresponding person and training a new model.

\begin{figure}[t]
	\begin{center}
		\includegraphics[width=\linewidth]{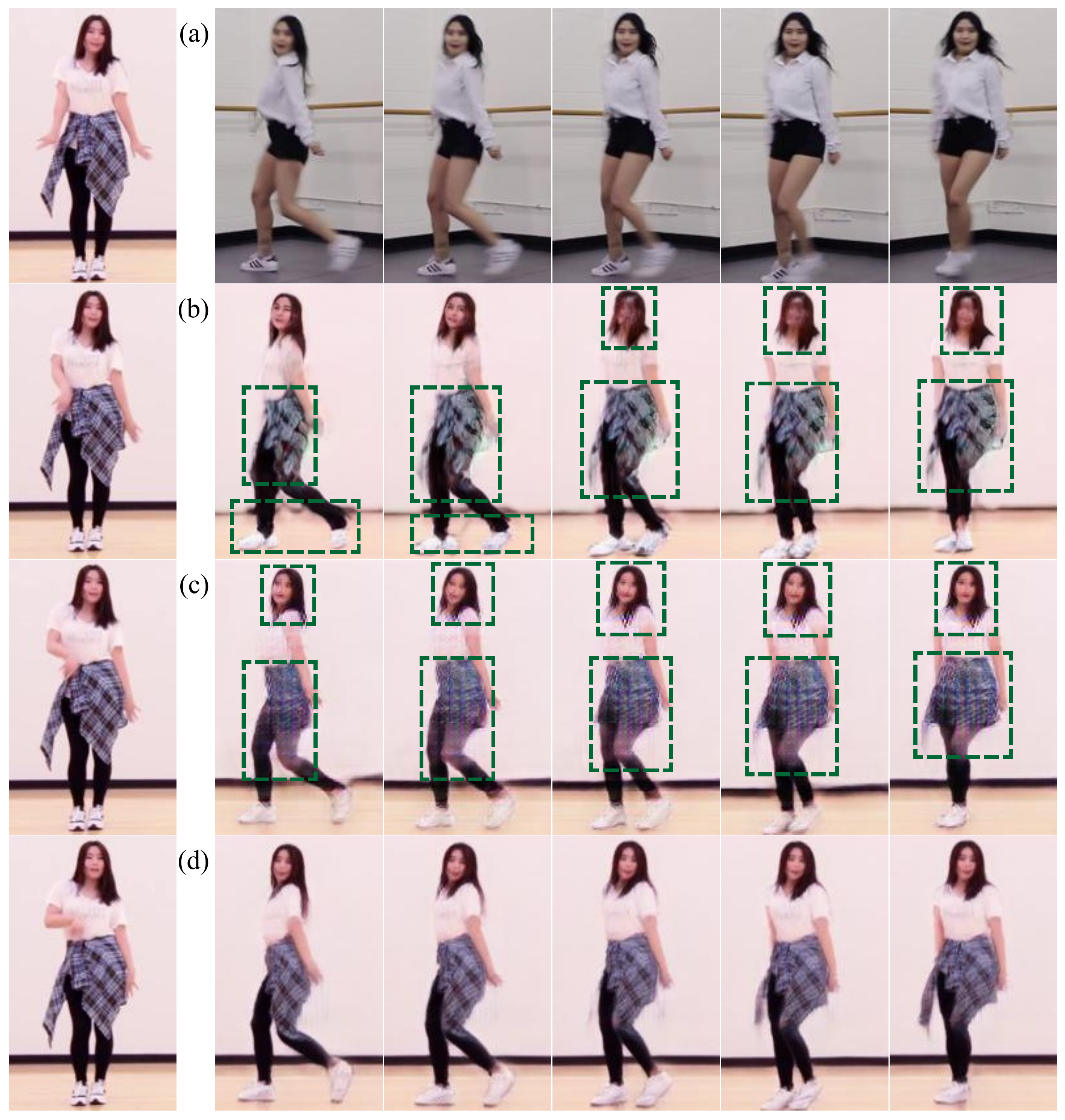}
	\end{center}
	\vspace{-10pt}
	\caption{The few-shot dancing video retargeting results of (b) Fs-vid2vid\cite{fsvid2vid} (c) MetaPix~\cite{metapix} and (d) our MetaDance, which retarget the five consecutive frames from the source dancing video in (a) to the target person shown in the first column. Only four consecutive frames of the target person in the first column are given for each method. The baseline models generate unsatisfactory results with clear artifacts and blurs as highlighted in green boxes. By contrast, MetaDance synthesizes photo-realistic results, which transfers the dance movements from the source video to the target person accurately and preserves the appearances of the target person.} 
	\vspace{-10pt}
	\label{fig:compare}
\end{figure}

To address the above limitations, some recent work~\cite{fsvid2vid,metapix} tackled dancing video retargeting in the regime of few-shot learning, which learns to synthesize videos of unseen persons by leveraging a few frames of these persons.
For example, MetaPix~\cite{metapix} utilized a first-order meta-learning algorithm Reptile~\cite{reptile} to learn a generic retargeting model, such that it can be fine-tuned for a specific person with a few frames. Fs-vid2vid~\cite{fsvid2vid} generated the network weights using the frames of the target person, to inject the appearance patterns of the target person into the synthesis.
However, the above work simply regarded a few frames of the target person as a batch of individual images without temporal correlations.
Specifically, Fs-vid2vid extracted the appearance features of each frame of the target person individually and aggregated them with an attention mechanism. MetaPix optimized the performance of each frame in parallel using meta-learning. They both generated unsatisfactory results with clear artifacts and blurs as shown in Fig~\ref{fig:compare}.

To address the above problems, this work proposes MetaDance for few-shot dancing video retargeting which synthesizes photo-realistic and temporally coherent videos for unseen persons with only a few frames.
Instead of regarding a few frames of a target person as a batch of images without temporal correlations as in previous work, we model these frames as a few dancing moves, where each dancing move contains two consecutive frames, to capture not only the appearance patterns but also the temporal dynamics of the person.
We further formulate the synthesis of a dancing video as the synthesis of a series of dancing moves, so that a video is synthesized in a move-by-move manner.
In this way, few-shot dancing video retargeting can be decomposed into synthesizing a series of dancing moves for a target person given a few dancing moves of this person, and we solve it with temporal-aware meta-learning.

Meta-learning trains a model on a variety of learning tasks, such that it can solve new tasks using only a small number of training samples. Different from previous meta-learning work, where each task is made up samples without temporal correlations such as a batch of images with various categories for image classification~\cite{sun2019meta,baik2020meta}, a task in our temporal-aware meta-learning is regarded as a temporal task and contains consecutive frames for video generation. 
Specifically, a temporal task first synthesizes a dancing sequence of a few consecutive frames through synthesizing a few dancing moves to tune a model for a specific target person. It then evaluates the image-level and temporal performances of the tuned model through generating another dancing sequence of the same person also in the move-by-move manner and further adjusts its parameters for initialization. 
In a task, tuning a model through synthesizing a series of dancing moves and evaluating the tuned model in terms of the visual quality and temporal stability through synthesizing other series of dancing moves enables the model to exploit not only the static appearance patterns in individual frames but also the dynamic movement characteristics among adjacent frames for effective adjustments of the parameters.
After training large numbers of temporal tasks, we obtain a generalized model, which can be efficiently tuned with a few frames in the move-by-move manner for an unseen person towards enhanced visual quality and strengthened temporal stability.

In practice, pairs of videos with two different persons performing the same movements are not available, thus we utilize poses of the person in the source video as the intermediate representations and synthesize a dancing video of a target person with a video of desired poses from a source video.
Specifically, our MetaDance includes a Temporal Dancing Generation Network (TDGN) to synthesize each dancing move. 
During meta-training, we sample a large variety of temporal tasks to meta-train the model TDGN. A temporal task is sampled from a dancing video of a same person, which consists of a reference frame $I_0$ and its pose $P_0$, a support dancing sequence to tune TDGN for a specific person, and a query dancing sequence to evaluate the tuned model and further adjust its parameters for initialization. Both the support and the query sequence contain consecutive frames and their poses. Given $I_0$, $P_0$ and the poses of the support sequence, we first tune TDGN to synthesize the support sequence, which is decomposed into the synthesis of a few dancing moves $\{{\mathcal{M}}_s\}$. For a dancing move ${\mathcal{M}}_s$, we synthesize two consecutive frames $\hat{I}_s$ and $\hat{I}_{s+1}$, where each frame is synthesized based on the  generated result of the previous frame. %the current frame $\hat{I}_s$ and then use it as the input to synthesize the next frame $\hat{I}_{s+1}$. 
We sum both the frame-level and temporal losses of all the dancing moves in the support sequence to update the parameters of TDGN from $\theta$ to ${\theta}^{\prime}$. We then use TDGN with parameters ${\theta}^{\prime}$ to synthesize the query dancing sequence through generating a few dancing moves. The frame-level and temporal losses of all the dancing moves in the query sequence are added up to in turn update the parameters $\theta$ of TDGN for initialization. After abundant temporal tasks are trained, we obtain a generalized model TDGN, which can be effectively adapted to unseen persons with only a few frames.

Due to the lack of a standard benchmark for few-shot dancing video retargeting, we further release a dataset named DeepDance, which consists of 200 dancing videos from YouTube. Each video is divided into short clips of continuous motions, which yields about 5000 clips. DeepDance provides human poses for each frame in the form of human body landmarks and densepose~\cite{densepose}.

Our work has three main contributions.
\begin{itemize}[]
	\item
	We propose a novel few-shot dancing video retargeting method named MetaDance, which synthesizes high-quality videos with temporal coherence for unseen persons with only a few frames. Through modeling a few frames of a target person as a series of dancing moves, where a dancing move contains two consecutive frames, we 
	could capture not only the appearance patterns but also the temporal dynamics of the person.
	\item
	We utilize temporal-aware meta-learning to obtain a generalized model through training various temporal tasks. Each temporal task learns how to utilize a few frames through the synthesis of dancing moves to exploit both the appearance features in individual frames and the dynamic characteristics among adjacent frames, such that it can synthesize a different dancing sequence with the improved visual quality and temporal stability. The meta-learned model can be efficiently tuned with a few frames in the move-by-move manner for an unseen person.
	\item
	We release a novel dancing video retargeting dataset named DeepDance and build a standard benchmark for evaluating few-shot dancing video retargeting. To our knowledge, DeepDance is currently the largest public dataset for dancing video retargeting.

\end{itemize}

\section{Related Work}
\subsection{Video Synthesis}
Video synthesis focuses on generating video content instead of static images, which needs to ensure the temporal consistency of the output videos compared with image synthesis. 
Existing video synthesis tasks can be divided into three categories, including unconditional video synthesis~\cite{2016generating,2019adversarial,mocogan,2017temporal}, future video prediction~\cite{2019predicting,2018folded,2017dual,2017decomposing,2020future,2018predicting,2019directional,2017flexible,2020transformation} and video-to-video synthesis~\cite{vid2vid,2017real,junting,2019deep,2019inpaint,2019vid2game,2019learning,2021real,2019deepflow,2021zooming,2021basic}. 

Unconditional video synthesis generates a video by mapping a sequence of random vectors to a sequence of video frames. For example, MoCoGAN~\cite{mocogan} decomposed the motion and content parts of a sequence and used a fixed latent code for the content and a series of latent codes to generate the motion. The synthesized videos are usually up to a few seconds on simple video content, such as facial motion.

Future video prediction aims to predict future video frames based on the past frames of the video. For example, some methods~\cite{2018predicting,2019directional} reduced the prediction space to high-level representations, such as human pose and semantic segmentation, which constitute good intermediate representations and are more tractable. Some approaches~\cite{2017flexible,2020transformation} assumed that visual information is already available in the past frames and explicitly modeled the transformations between the past frames and future frames for the prediction of future frames.

Video-to-video synthesis converts an input semantic video to an output video, which contains various tasks, such as transforming high-level representations to videos~\cite{vid2vid}, transferring the style of a reference image/video to an input video~\cite{2017real,2019learning,2021real},  filling in missing regions of a given video~\cite{2019inpaint,2019deepflow} and generating a video sequence with high-resolution from a low resolution input video~\cite{2021zooming,2021basic}.
Our work in dancing video retargeting belongs to the first category, where we utilize the poses of the person in the source video as intermediate high-level representations to synthesize a target video.

\subsection{Dancing Video Retargeting}
Dancing video retargeting aims to transfer the dance movements from a source video to a target person, thus synthesizing a video where the target person mimics the dancer in the source video, which falls into the category of video-to-video synthesis using poses as intermediate representations. Since a dancer in a video quickly changes postures with large movements, synthesizing a high-quality video with temporal coherence for a target person performing the same dance is extremely challenging.

Previous methods~\cite{vid2vid,everybody,dancegen,2021supervised} collected a several-minute-long video of a person with thousands of frames to train a personalized dancing retargeting model for this person. For example, EBDN~\cite{everybody} used 2 $\sim$ 4 minutes long video of a target person, which captures a sufficient range of motions of this person, to train a network for learning the mapping from the pose representations to images of the person. However, the trained model can only be used to generate videos of the same person.

Some recent work~\cite{fsvid2vid,metapix} tackled few-shot dancing video retargeting, which learns to synthesize videos of unseen persons using only a few frames of these persons.
For example, MetaPix~\cite{metapix} learned a retargeting model via a first-order meta-learning algorithm Reptile~\cite{reptile}, such that it can be fine-tuned for unseen persons with a few frames. Fs-vid2vid~\cite{fsvid2vid} injected the appearance features of the target person into the network through generating the network weights using a few frames of the target person.
However, these work simply regarded a few frames of the target person as a batch of images without temporal correlations and generated temporally incoherent dancing videos of low quality.
By contrast, our MetaDance models a few frames of a target person as a series of dancing moves, where each move contains two consecutive frames, to extract the appearance patterns as well as the dynamic characteristics of this person for high-quality video generation with temporal coherence.

\begin{figure*}[t]
	\begin{center}
		\includegraphics[width=\linewidth]{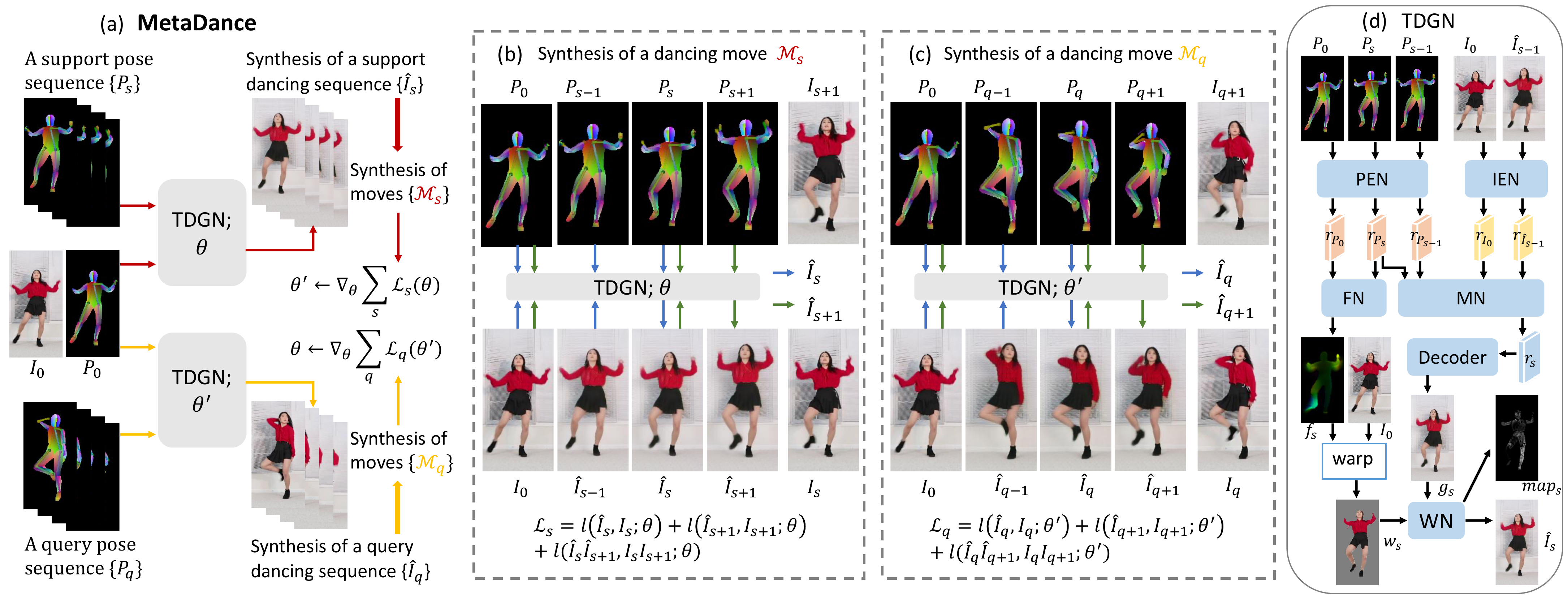}
	\end{center}
	\vspace{-10pt}
	\caption{(a). The overall framework of MetaDance during meta-training with a Temporal Dancing Generation Network (TDGN). A task is sampled from a dancing video, which consists of a reference frame $I_0$ and its pose $P_0$, a support dancing sequence and a query dancing sequence. Both the support and the query sequence contain consecutive frames and their poses. The red arrows tune TDGN to synthesize the support dancing sequence $\{\hat{I}_s\}$ through the synthesis of dancing moves $\{\mathcal{M}_s\}$, and update the parameters of TDGN from $\theta$ to ${\theta}^{\prime}$. The yellow arrows use the tuned TDGN with parameters ${\theta}^{\prime}$ to synthesize the query dancing sequence $\{\hat{I}_q\}$ through generating dancing moves $\{{\mathcal{M}}_q\}$, and in turn update the parameters $\theta$ of TDGN for initialization. (b). The pipeline of synthesizing a dancing move $\mathcal{M}_s$ in the support sequence and (c). the pipeline of synthesizing a dancing move ${\mathcal{M}}_q$ in the query sequence. The blue arrows and the green arrows generate two consecutive frames of a dancing move in order, where each frame is synthesized based on the generated result of the previous frame. (d). The architecture of the Temporal Dancing Generation Network (TDGN).}
	\vspace{-10pt}
	\label{fig:method}
\end{figure*}

\subsection{Meta-learning}
The goal of meta-learning is to train a model on a variety of tasks, such that it can solve new tasks using only a small number of training samples. 
Existing meta-learning methods can be generally divided into three categories, including optimization-based, metric-based and parameter prediction methods. 
In the first category, the optimization-based methods~\cite{finn2017model,li2017meta,nichol2018first,rajeswaran2019meta,baik2020meta,fu2019meta,banerjee2020meta} learned a a good initialization so that a few gradient updates on its parameters would lead to good performance on unseen tasks. In the second category, the metric-based methods ~\cite{vinyals2016matching,snell2017prototypical,sung2018learning,dong2018few,wang2019panet,wu2020meta,zhang2020sg,pambala2020sml} learned embeddings of training samples and test examples and a distance metric to measure the similarity between them. In the third category, the parameter prediction methods~\cite{bertinetto2016learning,shaban2017one,gidaris2019generating,gidaris2018dynamic,wang2019meta} learned to predict network parameters from a few annotated examples. 

Our work belongs to the first category, where we obtain a generalized model with good initialization parameters after training our model on a large number of tasks. However, previous meta-learning work only solved tasks that are made up of samples without temporal correlations and only optimized for image-level performance.
For example, Meta-SSD~\cite{fu2019meta} learned a good initialization for a detector to solve few-shot object detection through learning abundant tasks, where each task detects a few objects and optimizes the model with the classification and regression loss.
By contrast, we utilize temporal-aware meta-learning for few-shot dancing video retargeting, where each task contains consecutive frames for video generation and both the frame-level and temporal performances are optimized.

%------------------------------------------------------------------------
\section{Method}
Given a dancing video of a source person and a few frames of a target person, our goal is to generate a new video of the target person performing the same dance as the source video. Since corresponding pairs of videos containing two different persons performing the same motions are not available, we utilize poses of the person in the source video as intermediate representations. In this way, dancing video retargeting turns into synthesizing a dancing video of a target person given a video of desired poses extracted from a source video.
\subsection{Problem Setting}
We introduce the few-shot dancing video retargeting in the regime of mete-learning, which trains a set of learning tasks, denoted as $\mathcal{T}=\{{\mathcal{T}_i}\}$. Each task ${\mathcal{T}_i}$ is sampled from a dancing video $V_i$ in a $K$-shot setup following the setting of existing meta-learning work. For a task ${\mathcal{T}_i}$, we first sample $K$ consecutive frames with their poses from the video $V_i$ to tune the model for a specific person. We treat the first frame $I_0$ of the $K$ consecutive frames as the reference frame and its pose $P_0$ as the reference pose, and the remaining $K-1$ consecutive frames and their poses as a support sequence $S_i$.
We then sample additional consecutive frames and their poses from the video $V_i$ as the query sequence $Q_i$ to evaluate the performance of the tuned model and further adjust the parameters. There are some interval frames between the support sequence $S_i$ and the query sequence $Q_i$ to ensure that the two sequences consist of different movements.

The entire task set $\mathcal{T}$ is divided into two subsets without overlapping in terms of persons, including $\mathcal{T}^{\mathrm{seen}}$ and  $\mathcal{T}^{\mathrm{unseen}}$.  $\mathcal{T}^{\mathrm{seen}}$ is used for meta-training to train a generalized model, and  $\mathcal{T}^{\mathrm{unseen}}$ is used for meta-test to evaluate the performance of the meta-trained model in synthesizing dancing videos for unseen persons with only $K$ frames of a person.

In meta-test, we perform episodes to evaluate the performance of the meta-trained model following~\cite{gidaris2019generating}, where each episode randomly samples a task from $\mathcal{T}^{\mathrm{unseen}}$. Following the same setup in meta-training, each task is a $K$-shot setting associated with a reference frame $I_0$ and its pose $P_0$, and a support sequence $S_i$ with $K-1$ consecutive frames and their poses. A query sequence $Q_i$ is sampled with consecutive frames and their poses. %Given the support $S_i$, the model is tuned and then evaluated on the query set $Q_i$ to calculate the error. 
In a task sampled from $\mathcal{T}^{\mathrm{unseen}}$, a model is tuned to adapt to a specific person with the reference frame and its pose using a support sequence $S_i$, and is evaluated on a query sequence $Q_i$ to calculate the evaluation metrics.
We report the final results by averaging the results of all the episodes following previous meta-learning work~\cite{finn2017model}.

\subsection{Overview}
In order to solve few-shot dancing video retargeting, a model should possess two kinds of capabilities. On one hand, the model should effectively extract the appearance patterns of an unseen person with only a few frames, so as to synthesize photo-realistic results of this person in new poses. On the other hand, since the task involves video generation rather than image synthesis, the model needs to enforce the temporal coherence between the synthesized results to generate a temporally smooth video of the target person performing the same movements as the source video.

To solve the above issue, this work proposes MetaDance, which models a few frames of a target person as a few dancing moves and each dancing move contains two consecutive frames. Tuning a model for a target person with a few frames through synthesizing a few dancing moves helps the model capture the static appearance characteristics in individual frames as well as the temporal dynamics among the adjacent frames.
We further decompose the synthesis of a dancing video into the synthesis of a series of dancing moves, so that 
a video is generated in a move-by-move manner. In this way, few-shot dancing video retargeting turns into generating a series of dancing moves for a target person with a few dancing moves, and we adopt temporal-aware meta-learning to solve it. 
Temporal-aware meta-learning trains a large number of temporal task to obtain a generalized model, which can be efficiently tuned with a few frames in the move-by-move manner for an unseen person. We design a Temporal Dancing Generation Network (TDGN) to synthesize each move.

As shown in Fig.~\ref{fig:method} (a), a task in the meta-training stage of MetaDance contains the synthesis of a support dancing sequence depicted by the red arrows to tune the model TDGN for a target person, and a query dancing sequence depicted by the yellow arrows to evaluate the performance of the tuned model and further update the parameters for initialization. The support and the query sequence are sampled from a dancing video of a same person, each containing consecutive frames and their poses. There are some interval frames between the support sequence and the query sequence, to ensure that the two sequences consist of different movements.
First of all, the red arrows synthesize the support dancing sequence $\{\hat{I}_s\}$ through synthesizing dancing moves $\{\mathcal{M}_s\}$, where a move ${\mathcal{M}}_s$ is made up two consecutive frames $I_s$ and $I_{s+1}$, to exploit both the appearance features in individual frames and the dynamic characteristics among adjacent frames. As shown in Fig.~\ref{fig:method} (b), the synthesis of a dancing move ${\mathcal{M}}_s$ involves generating two consecutive frames $\hat{I}_s$ and $\hat{I}_{s+1}$ with the blue arrows and the green arrows in order, where each frame is synthesized based on the generated result of the previous frame. Both the image-level and the temporal losses of the generated move are calculated. After synthesizing all the dancing moves in the support sequence, the losses of each move are added up as the total loss of a support sequence to update the parameters of TDGN from $\theta$ to ${\theta}^{\prime}$ for this person.
Then the yellow arrows use the tuned TDGN with parameters ${\theta}^{\prime}$ to synthesize the query dancing sequence $\{\hat{I}_q\}$ through generating dancing moves $\{{\mathcal{M}}_q\}$. Fig.~\ref{fig:method} (c) describes the pipeline of synthesizing a dancing move ${\mathcal{M}}_q$ in the query sequence, where the blue arrows and the green arrows generate two consecutive frames $\hat{I}_q$ and $\hat{I}_{q+1}$ successively. Both the image-level and temporal losses of all dancing moves in the query sequence are added up to evaluate the ability of the tuned TDGN in synthesizing a different dancing sequence and in turn update the parameters $\theta$ of TDGN for initialization. %We will introduce the entire meta-learning stage of MetaDance in the following part.

As shown in Fig.~\ref{fig:method} (d), our model TDGN is made up of a pose extraction network (PEN) and an image extraction network (IEN) to extract features for the poses and images respectively, a flow network (FN) to predict the optical flows between the pose representations of the reference frame and the target frame for warping the reference frame, a modulating network (MN) to modulate the image representations and the pose representations, a decoder to render a rough image with the modulated representations, and a weighting network (WN) to predict the occlusion map between the rendered image and the warped image for the final synthesized result. We will introduce each component in Sec.~\ref{sec:network}.

\begin{algorithm*}[h] 
	\caption{Meta-training of MetaDance} 
	\label{alg:train} 
	\begin{algorithmic}[1] 
		\Require 
		$p(\mathcal{T})$: task distribution over the dancing videos.
		\Require
		$\theta$: parameters of the pre-trained TDGN.
		\Require
		$\alpha, \beta$: step size hyper-parameters.
		\Ensure
		${\theta}$: parameters of the meta-learned TDGN.
		\While{not done do}
		\State Sample batch of tasks $\mathcal{T}_i \sim p(\mathcal{T})$, where each task $\mathcal{T}_i$ is sampled from a dancing video. $\mathcal{T}_i$ consists of a reference frame $I_0$ and its pose $P_0$, a support sequence $S_i$ and a query sequence $Q_i$, both with consecutive frames and their poses.
		
		\For{all $\mathcal{T}_i$}
		\For{each dancing move ${\mathcal{M}}_{s} \sim  S_i$, which contains two consecutive frames and their poses}
	\State $\hat{I}_s = TDGN(I_0P_0,\hat{I}_{s-1}P_{s-1},P_s;\theta)$, where $\hat{I}_{s-1}$ is the result of the previous frame and $P_{s-1}$ is its pose.
	\State $\hat{I}_{s+1} = TDGN(I_0P_0,\hat{I}_{s}P_{s},P_{s+1};\theta)$.
	\State Calculate ${\mathcal{L}}_{s}=l(\hat{I}_{s},I_s)+l(\hat{I}_{s+1},I_{s+1})+l(\hat{I}_{s}\hat{I}_{s+1},I_{s}I_{s+1})$, where ($I_{s}$, $I_{s+1}$) are the ground-truth move.
	\EndFor
	\State Sum the losses of all dancing moves ${\mathcal{M}}_{s}$ in the support sequence $S_i$ and update ${\theta}_i^{\prime}= \theta-\alpha {\nabla}_{\theta}\sum_{s}{\mathcal{L}}_{s}({\mathcal{M}}_s;\theta)$.
	\For{each dancing move ${\mathcal{M}}_{q} \sim  Q_i$, which contains two consecutive frames and their poses}
	%with $\{I_0P_0,\hat{I}_{s-1}P_{s-1},I_sP_s,I_{s+1}P_{s+1}\}$, where $\hat{I}_{s-1}$ is the synthesized result of the previous frame}
\State $\hat{I}_q = TDGN(I_0P_0,\hat{I}_{q-1}P_{q-1},P_q;{\theta}^{\prime})$, where $\hat{I}_{q-1}$ is the result of the previous frame and $P_{q-1}$ is its pose. 
\State $\hat{I}_{q+1} = TDGN(I_0P_0,\hat{I}_{q}P_{q},P_{q+1};{\theta}^{\prime})$.
\State Calculate ${\mathcal{L}}_{q}=l(\hat{I}_{q},I_q)+l(\hat{I}_{q+1},I_{q+1})+l(\hat{I}_{q}\hat{I}_{q+1},I_{q}I_{q+1})$, where ($I_{q}$, $I_{q+1}$) are the ground-truth move.
\EndFor
\EndFor
\State Sum the losses of all dancing moves ${\mathcal{M}}_{q}$ in all query sequences $Q_i$ and update $\theta \leftarrow \theta-\beta {\nabla}_{\theta}{\sum}_{Q_i}\sum_{q}{\mathcal{L}}_{q}({\mathcal{M}}_q;{\theta}_i^{\prime})$.
\EndWhile
\end{algorithmic} 
\end{algorithm*}

\subsection{Temporal-aware Meta-learning}
We adopt temporal-aware meta-learning for few-shot dancing video retargeting, which trains a variety of temporal tasks to learn a generalized model that can be efficiently adapted to a new person with a few frames.
A temporal task consists of synthesizing a support dancing sequence to tune the model for a specific person and a query dancing sequence to evaluate the performance of the tuned model and further adjust its parameters for initialization. In this part, we first introduce how to synthesize a dancing sequence through generating a series of dancing moves, and then explain how to train a temporal task in meta-learning and how to solve an unseen temporal task in meta-test.
\subsubsection{The Synthesis of a Dancing Sequence}
Taking the synthesis of a support dancing sequence as an example in Fig.~\ref{fig:method} (a), given a reference frame $I_0$ and its pose $P_0$, the poses of a dancing sequence $\{P_s\}$, synthesizing a dancing sequence aims to generate consecutive frames as $\{\hat{I}_s\}$, which are not only individually photo-realistic but also temporally coherent among frames. 
We decompose the synthesis of a dancing sequence into the synthesis of a series of dancing moves denoted as $\{\mathcal{M}_s\}$, where each move contains two consecutive frames. The synthesis of a dancing move ${\mathcal{M}}_s$ involves generating two consecutive frames $\hat{I}_s$ and $\hat{I}_{s+1}$ successively depicted by the blue arrows and the green arrows in Fig.~\ref{fig:method} (b). Specifically, given a reference frame $I_0$ and its pose $P_0$, the poses of a dancing move $P_s$ and $P_{s+1}$, we synthesize a dancing move $\mathcal{M}_s$ = ($\hat{I}_s$, $\hat{I}_{s+1}$) as below:
\begin{equation}
\hat{I}_s = TDGN(I_0P_0,\hat{I}_{s-1}P_{s-1},P_s;\theta)
\end{equation}
\begin{equation}
\hat{I}_{s+1} = TDGN(I_0P_0,\hat{I}_{s}P_{s},P_{s+1};\theta)
\end{equation}
where $\hat{I}_{s-1}$ is the generated result of the previous frame and $P_{s-1}$ is its pose.

We use pixel-wise L1 loss and the perceptual loss~\cite{perceptual} as the image-level loss to encourage the visual similarity between the synthesized frame and the ground-truth frame as below:
\begin{equation}	
\mathcal{L}_{sl} = {||\hat{I}_{s}-I_s||}_1 + {||\hat{I}_{s+1}-I_{s+1}||}_1
\end{equation}
\begin{equation}	
\begin{split}	
\mathcal{L}_{sp} &= \sum_{m}~{||~{\phi}_m(\hat{I}_{s})-{\phi}_m(I_s)~||}_1 \\
&+ \sum_{m}~{||~{\phi}_m(\hat{I}_{s+1})-{\phi}_m(I_{s+1})~||}_1
\end{split}
\end{equation}
where ${\phi}_m$ indicates the $m$-th feature map in a VGG-19~\cite{vgg} network pre-trained on ImageNet~\cite{imagenet}. 

To enforce the temporal coherence between adjacent frames, we use a temporal discriminator $D_T$ to ensure that the results of consecutive frames resemble the temporal dynamics of the ground-truth consecutive frames given the same optical flows. The temporal loss is defined as GAN loss shown below:
\begin{equation}
\begin{split}
\mathcal{L}_{st} &= E[\text{log}D_T(I_s,I_{s+1},f_{s,s+1})] \\
&+ E[\text{log}(1-D_T(\hat{I}_{s},\hat{I}_{s+1},f_{s,s+1})]
\end{split}
\end{equation}
where $f_{s,s+1}$ denotes the optical flows between the frames $I_s$ and $I_{s+1}$ extracted from a pre-trained network~\cite{flownet}. 

To sum up, the loss $\mathcal{L}_s$ of a dancing move $\mathcal{M}_s$ is calculated as below:
\begin{equation}
\mathcal{L}_s=\lambda_l\mathcal{L}_{sl} + \lambda_p\mathcal{L}_{sp} + \lambda_t\mathcal{L}_{st}
\end{equation}

After synthesizing all the dancing moves in the sequence, we sum the loss of each move as the total loss of the dancing sequence, which assesses not only the visual quality of single frames but also the temporal stability of consecutive frames in this dancing sequence.

\subsubsection{Meta-training}
A temporal task in meta-training first synthesizes a support dancing sequence $\{\hat{I}_s\}$ through generating a few dancing moves $\{\mathcal{M}_s\}$, to tune the model for a specific person. Since the model generates dancing moves rather than individual frames and is optimized for both the image-level and the temporal performances, the tuned model could capture not only the appearance characteristics but also the temporal dynamics of the target person.
Specifically, given a reference frame $I_0$ and its pose $P_0$, and a pose dancing sequence $\{P_s\}$, it generates moves $\{\mathcal{M}_s\}$, as shown in the red arrows of Fig.~\ref{fig:method} (a). We sum the losses of all the dancing moves in the support sequence to update the parameters of TDGN from $\theta$ to ${\theta}^{\prime}$ as below:
\begin{equation}
{\theta}^{\prime}= \theta-\alpha {\nabla}_{\theta}\sum_{s}{\mathcal{L}}_{s}({\mathcal{M}}_s;\theta)
\end{equation}

After the parameters of TDGN is tuned with the support dancing sequence, a temporal task evaluates the image-level and temporal performances of the tuned model through generating a query dancing sequence, and further optimizes TDGN for initialization.
To be specific, the task generates moves $\{\mathcal{M}_q\}$ using TDGN with parameter ${\theta}^{\prime}$, taking the reference frame $I_0$ and its pose $P_0$, and a pose sequence $\{P_q\}$ as inputs, as shown in the yellow arrows of Fig.~\ref{fig:method} (a). We sum the image-level and temporal losses of all the dancing moves in the query sequence and calculate the gradients of the initial $\theta$ to optimize $\theta$ as below:
%Instead of calculating the gradient of ${\theta}^{\prime}$ to optimize ${\theta}^{\prime}$, we 
\begin{equation}
\theta \leftarrow \theta-\beta {\nabla}_{\theta}\sum_{q}{\mathcal{L}}_{q}({\mathcal{M}}_q;{\theta}^{\prime})
\end{equation}

After training a large number of temporal tasks, where each task learns how to utilize a few frames to extract the visual features and the dynamic characteristics of a target person to synthesize a different dancing sequence with the improved visual quality and temporal stability, we obtain a generalized model that can be efficiently tuned with a few frames in the move-by-move manner for an unseen person.

The entire meta-training stage of MetaDance is summarized in Alg.~\ref{alg:train}. In each task, a reference frame and its pose, a support dancing sequence and a query dancing sequence from a dancing video are sampled. Line 4-8 synthesizes the support dancing sequence through generating a series of dancing moves, and line 9 tunes the model TDGN to adapt to a specific person. Line 10-14 synthesizes the query dancing sequence with the tuned model through generating a series of dancing moves, and line 16 in turn uses the image-level and temporal losses of all the query sequences in a batch of tasks to optimize the parameters of the model for initialization.

\subsubsection{Meta-test}
In the meta-test stage, each temporal task is sampled from an unseen person to evaluate the performance of the meta-trained model. The steps are the same as the meta-training stage, which first uses the support sequence to tune the model for a specific person through generating a series of dancing moves and then synthesizes the query dancing sequence with the tuned model. The only difference from the meta-training stage is that the results of the synthesized query sequence will be directly evaluated without a backward pass to optimize the parameters.

\subsection{Temporal Dancing Generation Network (TDGN)}\label{sec:network}
As shown in Fig.~\ref{fig:method} (d), our model TDGN is used to synthesize dancing moves and is made up of six components including a pose extraction network (PEN), an image extraction network (IEN), a flow network (FN), a modulating network (MN), a decoder and a weighting network (WN).
To synthesize a dancing move ${\mathcal{M}}_s$, which contains two consecutive frames ($\hat{I}_s$,$\hat{I}_{s+1}$), TDGN first synthesizes $\hat{I}_s$ of pose $P_s$ with the reference frame $I_0$ and its pose $P_0$, the pose of its previous frame $P_{s-1}$ and the generated result ${\hat{I}_{s-1}}$ using five steps as below. \textbf{In the first step}, PEN and IEN extract pyramidal features for the poses and images respectively. \textbf{In the second step}, the pose presentations of the reference frame $r_{P_0}$ and the target frame $r_{P_s}$ are fed into FN to predict the optical flows $f_s$, which are used to warp the foreground of the reference frame $I_0$ to the warped image $w_s$. \textbf{In the third step}, MN modulates the image representations of the reference frame $r_{I_0}$ and the previously generated result $r_{{\hat{I}_{s-1}}}$ with the pose representations of the target frame $r_{P_s}$ and the previously generated result $P_{s-1}$. \textbf{In the fourth step}, the modulated representations $r_s$ are fed into a decoder to generate a rough image $g_s$. \textbf{In the fifth step}, the warped image $w_s$ and the rough image $g_s$ are fed into WN to predict the occlusion map ${map}_s$ for the final synthesized result as $\hat{I}_s = {\text{map}}_s \odot w_s + (1 - {\text{map}}_s) \odot g_s$. TDGN then synthesizes $\hat{I}_{s+1}$ of pose $P_{s+1}$ with the reference frame $I_0$ and its pose $P_0$, the pose of its previous frame $P_{s}$ and the generated result ${\hat{I}_{s}}$ using the same five steps. In the following part, we will introduce the details of each component.

\subsubsection{Pose Extraction Network (PEN) and Image Extraction Network (IEN)}
Both PEN and IEN extract pyramid features from $N$ levels for the poses and images respectively with the feature pyramid network (FPN)~\cite{fpn}. Each FPN contains $N$ stages and each stage composes of a downsample convolution with a stride of 2 followed by two residual blocks~\cite{resnet}.

\subsubsection{Flow Network (FN)}
FN predicts the optical flows between the pose representations of the reference frame and the target frame, which adopts the structure of the appearance flow estimation network~\cite{PFAFN}. It learns to generate coarse optical flows at each pyramid level and refine the flows in the next level. The predicted optical flows are used to warp the foreground of the reference frame for the warped image, which preserves the appearance details of the target person.

\subsubsection{Modulating Network (MN)}
MN modulates the image features using the pose representations to effectively preserve the semantic layout of the target pose. Since both the pose features and the image features are pyramidal with $N$ levels, MN contains $N$ modulating layers, where each layer adopts the SPADE~\cite{spade} structure. The images features and the pose features at level-$l$ are fed into the $l$-th layer of MN to obtain the modulated features.

\subsubsection{Decoder}
The decoder takes the modulated pyramid features after MN as inputs to render a rough image, which synthesize the person of the reference frame with the target pose. The decoder is made up of an upsampling layer followed by two residual block with skip connections. 

\subsubsection{Weighting Network (WN)}
WN predicts the occlusion map, which weights between the warped image from FN and the rough image from the decoder to combine them for the final result. The warped image preserves the appearance details of the target person, but is not completely reliable since the large movements between the reference pose and the target pose often lead to self-occlusion. The generated image from the decoder is generally qualified, but sometimes loses details of the target person. The combination of the warped image and the generated image produces a more realistic result with the appearance details preserved. WN adopt the Res-UNet~\cite{resunet}, which is built upon a UNet~\cite{unet} architecture and combines with residual connections.

\begin{table*}\small\centering
\vspace{10pt}
\caption{Image-level evaluation of few-shot dancing video retargeting, where a few frames of a target person and a source pose sequence is sampled from a same video in a task. The evaluation metrics are Mean Square Error (MSE), Peak Signal-to-Noise Ratio (PSNR) and Structural Similarity (SSIM), which are averaged over 400 episodes. Smaller valuer of MSE and higher scores of PSNR and SSIM indicate higher visual quality of the synthesized images. The best performance is bold. } 
\scalebox{1.0}{
\begin{tabular}{c|ccc|ccc|ccc|ccc}
	\toprule[1.5pt]
	\multicolumn{1}{c|}{}&\multicolumn{3}{c|}{Shot-3}&\multicolumn{3}{c|}{Shot-5}&\multicolumn{3}{c|}{Shot-8}&\multicolumn{3}{c}{Shot-10}\\
	\toprule[1.5pt]
	\multirow{1}*{Metric}&MSE&PSNR&SSIM&MSE&PSNR&SSIM&MSE&PSNR&SSIM&MSE&PSNR&SSIM\\
	\hline
	\multirow{1}*{PrePix}&2551.84&14.52&0.448&2537.93&14.56&0.443&2440.35&14.75&0.458&2394.36&14.77&0.455\\
	\multirow{1}*{MetaPix}&2131.94&15.29&0.492&2058.10&15.44&0.497&1996.91&15.53&0.502&2031.39&15.51&0.504\\
	\multirow{1}*{Fs-vid2vid}&2335.03&14.95&0.469&2298.55&15.13&0.482&2217.14&15.28&0.502&2141.36&15.42&0.506\\
	\toprule[1.5pt]
	%\multirow{1}*{PreDance}&&&&&&&&&&&&\\
	\multirow{1}*{MetaDance}&\textbf{1690.85}&\textbf{16.43}&\textbf{0.551}&\textbf{1654.24}&\textbf{16.61}&\textbf{0.555}&\textbf{1653.44}&\textbf{16.63}&\textbf{0.564}&\textbf{1612.18}&\textbf{16.74}&\textbf{0.567}\\
	\bottomrule[1.5pt]
\end{tabular}}
\vspace{-10pt}
\label{tab:overall}
\end{table*}

\section{Experiments}
\subsection{Dataset}
At present, there is no standard benchmark for few-shot dancing video retargeting. In this work, we release a dataset named DeepDance, which consists of 200 dancing videos collected from YouTube.
For each video, we first extract human body landmarks and densepose~\cite{densepose, Ge_2021_CVPR} for each frame. Then we remove the invalid frames including frames where no person appears or a person appears without full body, frames containing more than one person that overlaps with others, and frames where no motion is detected more than a certain number of frames.  We further divide the videos into into short clips of continuous motions, which yields about 5000 clips. The entire dataset is split into a training set, which contains 180 videos with 4463 clips, and a test set, which contains 20 videos with 360 clips.

\subsection{Implementation Details}
We crop the person region of each frame and resize it to 128 $\times$ 256. We concatenate the human body landmarks and densepose of each frame following vid2vid~\cite{vid2vid} as the input pose.  We first pretrain the model TDGN on the training set of DeepDance for 60K iterations, with a batch size of 16 distributed over 8 GPUs. The learning rate is 0.0001 at first and decays linearly from 30K iterations. 
For the meta-training stage, a total of 300K tasks are sampled and each task is trained using 3 gradients steps with learning rate $\alpha=0.0001$ and $\beta=0.00005$.

During meta-test, we sample 20 random episodes for each unseen person in the test set of DeepDance and a total of 400 episodes are constructed. Specifically, each episode samples a $K$-shot task with $K$ consecutive frames to tune the model and another 50 query frames to evaluate the tuned model. Results over 400 episodes are averaged as the final evaluation results.

\subsection{Baseline Models}
We adopt the models in previous few-shot dancing video retargeting to serve as three baseline models.
\subsubsection{PrePix~\cite{metapix}} It learns a generative model on the training set of DeepDance and adapts the model to a few frames of a target person through fine-tuning.

\subsubsection{MetaPix~\cite{metapix}} It utilizes a first-order meta-learning algorithm, Reptile~\cite{reptile} to meta-optimize a generative adversarial network, such that a meta-learned model can be specialized to a few frames of a target person.

\subsubsection{Fs-vid2vid~\cite{fsvid2vid}} It uses a few frames of a target person to dynamically configure the video synthesis mechanism via generating the network weights using the provided frames.

\subsection{Quantitative Results}
\begin{table}\small\centering
\vspace{0pt}
\caption{Image-level evaluation of few-shot dancing video retargeting, where a task samples a few frames of a target person and samples a source pose sequence of another person. The evaluation metric is Fr\'{e}chet Inception Distance (FID), which is averaged over 400 episodes and smaller valuer indicates higher visual quality of the synthesized images. Results on different shots are further averaged as the mean value. The best performance is bold.} 
\scalebox{1.0}{
\begin{tabular}{c|ccccc}
	\toprule[1.5pt]
	
	\multirow{1}*{}&Shot-3&Shot-5&Shot-8&Shot-10&Mean\\
	\hline
	\multirow{1}*{PrePix}&157.37&158.44&159.76&155.36&157.73\\
	\multirow{1}*{MetaPix}&148.14&151.44&149.36&147.82&149.19\\
	\multirow{1}*{Fs-vid2vid}&151.72&153.94&152.78&152.31&152.69\\
	\toprule[1.5pt]
	\multirow{1}*{MetaDance}&\textbf{144.17}&\textbf{146.99}&\textbf{145.29}&\textbf{143.77}&\textbf{145.06}\\
	\bottomrule[1.5pt]
\end{tabular}}
\vspace{-10pt}
\label{tab:fid}
\end{table}

\subsubsection{Evaluation Metric} We use several evaluation metrics to evaluate both the image-level and temporal performance.

\paragraph{MSE} Mean Square Error (MSE) measures the  difference between the synthesized images and the ground-truth images in pixel level, and \textbf{lower} MSE means better results.

\paragraph{PSNR} Peak Signal-to-Noise Ratio (PSNR) provides an objective criterion to measure the distortion or noise level of the generated images compared with the ground-truth images, where \textbf{higher} scores of PSNR are better.

\paragraph{SSIM~\cite{sun2019meta}} Structural Similarity (SSIM) calculates the similarity of the synthesized images and the ground-truth images based on luminance, contrast and structural similarity. \textbf{Lager} SSIM means higher quality.

\paragraph{FID~\cite{fid}} Fr\'{e}chet Inception Distance (FID) measures the distance between the distributions of real
data and generated data, which is commonly used to quantify the fidelity of the synthesized images. This metric is adopted when a task in meta-test samples a few frames of a target person and samples a source dancing sequence of another person, thus no ground-truth target sequence is available. \textbf{Lower} score of FID indicates higher quality of the results.

\paragraph{TWE~\cite{blind}} We follow previous work~\cite{blind,blind2} to adopt Temporal Warping Error (TWE) to evaluate the temporal consistency between the synthesized results based on the flow warping error. \textbf{Lower} scores of TWE means higher temporal coherence among synthesized results.

\begin{table}\small\centering
\vspace{0pt}
\caption{Temporal evaluation of few-shot dancing video retargeting. The evaluation metric is Temporal Warping Error (TWE), which is averaged over 400 episodes and smaller valuer indicates higher temporal coherence. Results on different shots are further averaged as the mean value. The best performance is bold. Compared with the baseline models, the mean temporal warping error on MetaDance decreases by 66.6\%, 46.6\%, 42.4\% respectively.} 
\scalebox{1.0}{
\begin{tabular}{c|ccccc}
	\toprule[1.5pt]
	
	\multirow{1}*{}&Shot-3&Shot-5&Shot-8&Shot-10&Mean\\
	\hline
	\multirow{1}*{PrePix}&665.28&673.03&647.63&655.92&660.47\\
	\multirow{1}*{MetaPix}&422.46&418.57&407.70&405.48&413.55\\
	\multirow{1}*{Fs-vid2vid}&435.23&402.15&348.63&345.09&382.78\\
	\toprule[1.5pt]
	\multirow{1}*{MetaDance}&\textbf{233.17}&\textbf{227.68}&\textbf{211.22}&\textbf{210.53}&\textbf{220.65}\\
	\bottomrule[1.5pt]
\end{tabular}}
\vspace{-10pt}
\label{tab:temp}
\end{table}

\begin{figure*}[t]
\begin{center}
\includegraphics[width=\linewidth]{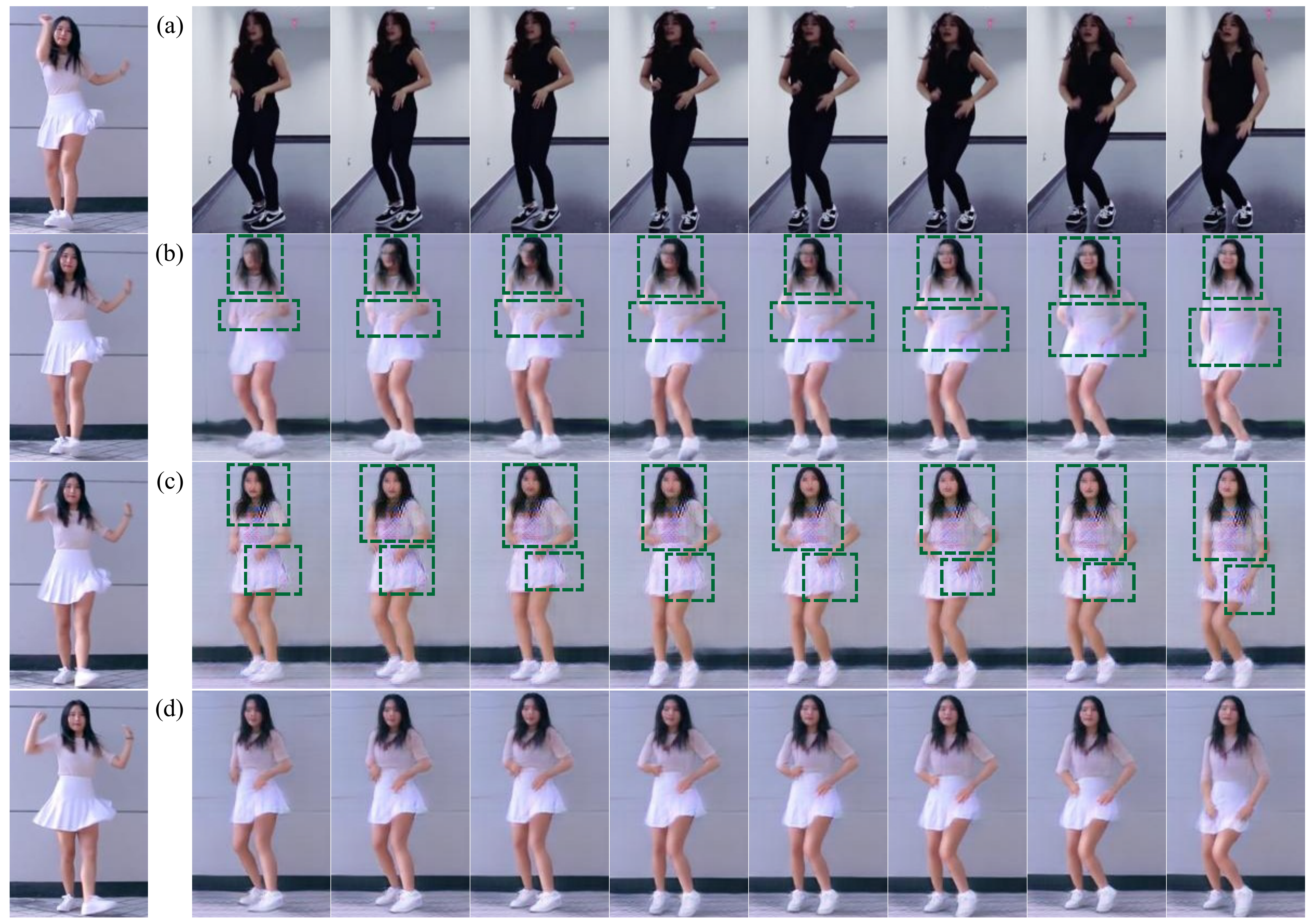}
\end{center}
\vspace{-10pt}
\caption{Visual comparison of few-shot dancing video retargeting, which transfers the dance movements from the source dancing sequence in (a) with eight consecutive frames to the target person shown in the first column. Only four consecutive frames of the target person in the first column are provided for each method. The synthesized sequences of (b) Fs-vid2vid~\cite{vid2vid}, (c) MetaPix~\cite{metapix}, and (d) MetaDance are respectively shown in the second, third and fourth row. The baseline models generate unsatisfactory results with clear artifacts and blurs as highlighted in green boxes. By contrast, our model generates high-quality results, which transfer the dancing movements of the source sequence to the target person accurately and preserves the appearances of the target person including hair, face and clothes.} 
\vspace{-10pt}
\label{fig:results}
\end{figure*}

\subsubsection{Comparison with Baselines}
We evaluate the results of the few-shot dancing video retargeting quantitatively in terms of both the image-level and temporal performance as below:

\paragraph{Image-level Comparison}
We adopt MSE, PSNR and SSIM as the evaluation metrics to assess the image-level visual quality of the synthesized results as shown in Tab.~\ref{tab:overall}, where a task samples a few frames of a target person and a source pose sequence from a same video so that the ground-truth target sequence is available for evaluation. Our MetaDance outperforms the baseline models by a large margin on all the evaluation metrics with different shots. Given any number of frames of a target person, MetaDance can efficiently extract appearance patterns of the target person and synthesize dancing sequences that preserve the appearance characteristics of the person with high visual quality. 
We further observe that each indicator of the models improves generally when the number of the frames of the target person increases. More frames of a target person show more diversified poses of this person, so that a model can better capture the appearance features of the target person without overfitting to some specific poses, which accounts for the enhanced performance of lager shots. Under all shots in the experiments, MetaDance exceeds the baseline models, showing its superiority in synthesizing high-quality results for unseen persons in new dancing poses with the appearance characteristics preserved.

We also evaluate the image-level performance of the models when a task samples a few frames of a target person and samples a source pose sequence of another person, with FID as the evaluation metric. As Tab.~\ref{tab:fid} shows, compared with the baseline models, our MetaDance achieves the best performance on all the different shots, which synthesizes dancing sequences with higher fidelity. After tuning our model with only a few frames of a target person, given an arbitrary dancing video, it generates a photo-realistic video of the target person performing the same movements as the source video. 
When comparing the performance of different shots, FID does not decreases with the increasing numbers of the frames of the target person, which shows that more frames for tuning a model dose not improve the fidelity of the synthesized results. Given any number of frames of a target person, MetaDance produces more realistic videos for this person performing various dancing movements.

\begin{figure*}[t]
\begin{center}
\includegraphics[width=\linewidth]{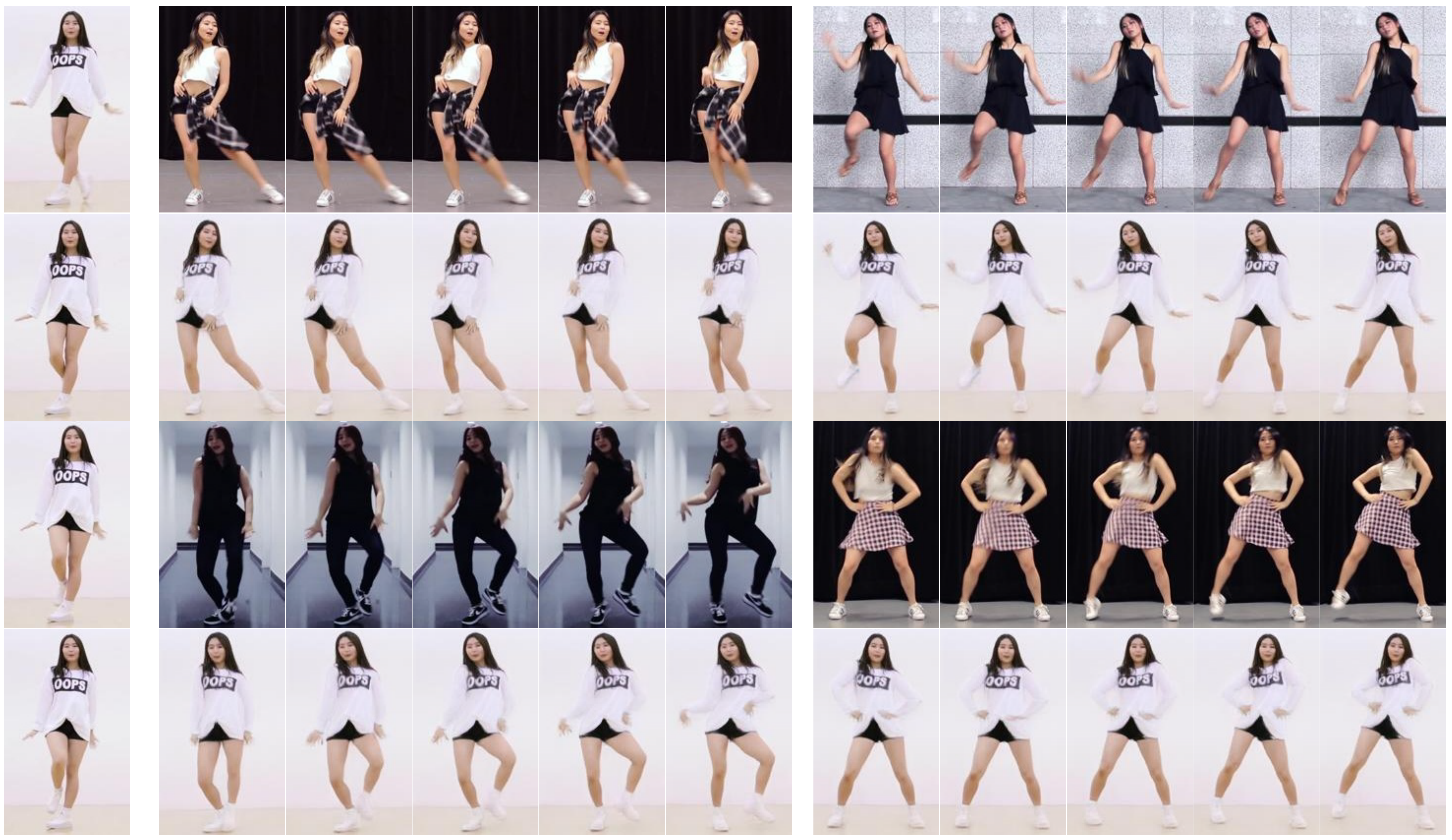}
\end{center}
\vspace{-10pt}
\caption{The visual results of MetaDance in transferring different dancing sequences of different persons to a target person. The first column shows the target person, where only four consecutive frames of this person are provided. The source dancing video sequences are shown in the first and the third row and the synthesized sequences for the target person are shown in the second and the fourth row. Given only a few frames of the target person, our model can synthesize high-quality video sequences where the target person performs various dancing moves naturally.} 
\vspace{-10pt}
\label{fig:more}
\end{figure*}

\paragraph{Temporal Comparison}
To evaluate the temporal coherence of the generated frames, we adopt TWE following previous work~\cite{blind,blind2} and list the results in Tab.~\ref{tab:temp}. Compared with the baseline methods, our MetaDance achieves significantly higher temporal consistence of all different shots, indicating that our method can synthesize exceedingly more temporally smooth dancing sequences, given any number of frames of a target person. 
We further average the temporal warping error of different shots, and the mean error of MetaDance decreases by 66.6\%, 46.6\%, 42.4\% than PrePix, MetaPix and Fs-vid2vid respectively. The large superiority of our method in terms of the temporal stability demonstrates the efficiency of our temporal-aware meta-learning in synthesizing temporal coherent results. Different from the baseline models that regard a few frames of a target person as a batch of images without any temporal correlations, we model these frames as a few dancing moves, where a dancing move contains two consecutive frames, to capture not only the appearance patterns but also the temporal dynamics of the target person. Through training a variety of temporal tasks, we learn a generalized model that can be effectively adapted to an unseen person with temporal coherence using only a few moves. 

We further analyze the influence of different shots in temporal coherence and find that the models synthesize temporally more stable results with more frames except for the baseline model PrePix. When tuning a model with more frames, MetaPix and Fs-vid2vid tend to smooth the generated results since the performance of a lager batch is optimized while MetaDance captures richer temporal dynamics of the target person. The overwhelming performance of MetaDance than the baselines proves the effectiveness of the temporal-aware meta-learning in solving few-shot dancing video retargeting.

\subsection{Qualitative Results}
We perform visual comparison of our MetaDance and the baseline models Fs-vid2vid~\cite{fsvid2vid} and MetaPix~\cite{metapix} in Fig.~\ref{fig:results}, which synthesizes a dancing sequence for a target person through transferring the dancing movements of the source sequence to the target person. The first column show the target person, where only four consecutive frames are provided for each method.
As shown in the second row of Fig.~\ref{fig:results}, Fs-vid2vid generates unsatisfactory results with clear artifacts. For example, the faces, arms, and clothes are messy in the synthesized sequence. It can not accurately transfer the movements of the source sequences to the target person and synthesize realistic results of the target person in corresponding poses. Fs-vid2vid solves few-shot video retargeting through predicting the weights of the network using the frames of the target person to inject the appearance patterns into the synthesis. Such a method only helps preserve the appearance characteristics of the target person, but can not solve the generation of an unseen person in different poses.
The third row of Fig.~\ref{fig:results} shows that MetaPix generates a low-quality dancing sequence with serious blurring. MetaPix fails to synthesize realistic face and clothes, thus it can not preserve the appearances of the target person. It adopts a simple pose-to-image translation framework to learn a mapping between the poses and the person images. When only a few frames are provided, MetaPix can not adjust its parameters effectively for the target person to preserve the appearance characteristics in the synthesized results.
By contrast, our MetaDance synthesizes a high-quality dancing sequence, which not only transfers the dancing movements of the source video sequence to the target person accurately, but also preserves the appearances of the target person including hair, face and clothes. 

We further show the video clips for comparison in \url{https://github.com/geyuying/MetaDance}. Fs-vid2vid and MetaPix generate non-smooth videos with noticeable flickering while MetaDance synthesizes more temporally coherent videos with high visual quality.
Through training various temporal tasks, where each task models a few frames of a target person as a series of dancing moves to exploit the appearance patterns and the temporal dynamics of the person, we learn a generalized model that can be efficiently tuned with a few frames and synthesize photo-realistic videos with temporal consistency.

We present the visual results of MetaDance in transferring various dancing sequences of different persons to a target person in Fig.~\ref{fig:more}. The dancing sequences are presented in the first and the third row, which cover diverse movements including swinging the leg, moving the arms, and swaying the shoulders. Given only four consecutive frames of the target person shown in the first column, our MetaDance synthesizes photo-realistic dancing sequences, where the target person performs various dancing moves naturally. MetaDance learns a generalized video retargeting model that can 
synthesize satisfactory video sequences for a target person performing various dancing moves. We also show a video clip of MetaDance in \url{https://github.com/geyuying/MetaDance}, which transfers the same dance movements to different target persons given four consecutive frames of each target person. MetaDance generates temporally smooth dancing videos with satisfactory visual quality using only a few frames of a target person.

\subsection{Ablation study}

\subsubsection{The Number of Dancing Moves} Our MetaDance trains a variety of temporal tasks, where each task tunes the model for a specific person through synthesizing a few support moves, and then synthesizes a few query moves with the tuned model to further update the model parameters. We perform an ablation study to explore the effects of the number of dancing moves in meta-training.
As shown in Tab.~\ref{tab:moves}, the model achieves better results generally on all the evaluation metrics including MSE, PSNR, SSIM, FID for image-level evaluation and TWE for temporal evaluation with the increasing number of dancing moves in meta-training. We can summarize that more dancing moves in temporal tasks for meta-training a model contribute to generating results with higher visual quality and stronger temporal consistency during meta-test.
Tuning a model with more support dancing moves and evaluating the tuned model on more query dancing moves during meta-training forces the model to exploit the appearance features and the temporal dynamics of a target person more effectively to improve its frame-level and temporal performances in synthesizing a different dancing sequence.

\begin{table}\small\centering
\vspace{0pt}
\caption{Ablation study on the effects of the number of dancing moves in meta-training. Both the image-level and temporal performance are evaluated with 8 shots in meta-test. The best performance is bold.} 
\scalebox{1.0}{
\begin{tabular}{c|ccccc}
	\toprule[1.5pt]
	
	\multirow{1}*{Moves}&MSE&PSNR&SSIM&FID&TWE\\
	\hline
	\multirow{1}*{1}&1737.19 &16.47&0.555&145.44&224.04\\
	\multirow{1}*{2}&1660.32 &16.62 &0.561&145.14&215.93\\
	\multirow{1}*{3}&1653.44 &16.63 &0.564&145.29&211.22\\
	\multirow{1}*{4}&\textbf{1640.60} &\textbf{16.67} &\textbf{0.567}&\textbf{144.63}&\textbf{209.79}\\
	\bottomrule[1.5pt]
\end{tabular}}
\vspace{-10pt}
\label{tab:moves}
\end{table}

\subsubsection{Utilization of a Few Frames} Few-shot dancing video retargeting aims to synthesize videos for unseen persons by leveraging a few frames of them. We perform an ablation study to analyze different ways to utilize the given frames with different methods.

\paragraph{PreFrame and MetaFrame} They regard a few frames as a batch of individual images and adopt an image synthesis method to synthesize each frame with a model similar to our model TDGN, except that the inputs do not include the pose and the synthesized result of the previous frame. PreFrame pretrain a model on the training set of DeepDance while MetaFrame meta-trains a model using MAML~\cite{finn2017model}. They both tune the model with a few frames and synthesize the target sequence in a frame-by-frame manner. 

\paragraph{PreMove and TD-free} They regard a few frames as a series of moves like MetaDance, where a move contains two consecutive frames. While PreMove directly pretrains a model with the training set of DeepDance, TD-free meta-trains a model with a large number of tasks, which adopts the same method as MetaDance except that it does not use a temporal discriminator to enforce the temporal loss on each dancing move. They both tune the model with a few frames and synthesize the target sequence in a move-by-move manner.

\begin{table}\small\centering
\vspace{5pt}
\caption{Ablation study on different ways to utilize a few frames of a target person. Both the image-level and temporal performance are evaluated with 5 shots in meta-test. The best performance is bold.} 
\scalebox{1.0}{
\begin{tabular}{c|ccccc}
	\toprule[1.5pt]
	
	\multirow{1}*{}&MSE&PSNR&SSIM&FID&TWE\\
	\hline
	\multirow{1}*{PreFrame}&1971.37&15.84&0.529&149.24&299.30\\
	\multirow{1}*{MetaFrame}&1696.63&16.48&0.545&148.38&248.82\\
	\multirow{1}*{PreMove}&1873.32&16.10&0.537&148.60&293.18
	\\
	\multirow{1}*{TD-free}&1726.22&16.48&0.550&147.86&245.87\\
	\multirow{1}*{MetaDance}&\textbf{1654.24}&\textbf{16.61}&\textbf{0.555}&\textbf{146.99}&\textbf{227.68}\\
	\bottomrule[1.5pt]
\end{tabular}}
\vspace{-10pt}
\label{tab:ablation}
\end{table}

\paragraph{Results} The results of different models which utilize a few frames of a target person in different ways are shown in Tab.~\ref{tab:ablation} and we have the following observations. First of all, PreFrame performs the worst of all methods in terms of both the image-level and temporal evaluation. Treating a few frames as individual images and applying an image synthesis method to synthesize each frame separately ignore the temporal correlations among frames, thus generating incoherent videos of low visual quality. Compared with PreFrame, PreMove achieves better performance in the visual quality and the temporal stability. Tuning a model with a few frames through generating a series of moves helps capture the appearance patterns and temporal dynamics of the target person, which facilitates the video generation with temporal smoothness.

Second, MetaFrame performs better than PreFrame, and MetaDance shows enhanced results over PreMove in the image-level and temporal evaluation. MetaFrame and MetaDance both adopt a meta-learning method, which trains large numbers of tasks, thus their meta-trained models can be more efficiently adapted to unseen persons with a few frames. MetaDance further outperforms MetaFrame on all the evaluation metrics. While each task in MetaFrame learns to extract the appearance features of a person through synthesizing a batch of individual images, each temporal task in MetaDance learns to exploit dynamic characteristics besides visual information through generating a series of moves. The superiority of MetaDance over MetaFrame demonstrates the effectiveness of our temporal-aware meta-learning.

Finally, the image-level and temporal performance of TD-free drop compared with MetaDance. TD-free adopts the same method as MetaDance except that it does not use a temporal discriminator to exert the temporal loss on the synthesis of dancing moves. On one hand, a temporal discriminator can effectively enforce the temporal coherence between the adjacent frames to generate more temporally coherent videos since MetaDance achieves smaller temporal warping error than TD-free. On the other hand, strengthening the temporal coherence of the synthesized videos contributes to higher visual quality as MetaDance also performs better than TD-free on the image-level evaluation metrics.

%------------------------------------------------------------------------
\section{Conclusion}

In this work, we propose a novel few-shot dancing video retargeting method named MetaDance, which synthesizes photo-realistic and temporally coherent videos for unseen persons with only a few frames. %We formulate few-shot dancing video retargeting as synthesizing a series of dancing moves for a target person with a few dancing moves of this person, where a move contains two consecutive frames. 
We utilize temporal-aware meta-learning to train a variety of temporal tasks for a generalized model, where each task models a few frames of a target person as a series of dancing moves to exploit the appearance patterns and the temporal dynamics of the person. The meta-trained model can be efficiently adapted to new target persons using a few frames in the move-by-move manner.
Extensive evaluations clearly show that MetaDance outperforms the baseline models by a large margin in terms of the visual quality and the temporal stability. 

\bibliographystyle{IEEEtran}
\bibliography{IEEEfull.bib}

\end{document}